\newif\ifconfver

\newif\ifonecoltab
\onecoltabtrue        

\newif\ifplainver  
\plainvertrue

\ifplainver
    \confverfalse                         
\fi

\ifconfver
     \documentclass[10pt,journal]{IEEEtran}
\else
    \ifplainver
        \documentclass[11pt]{article}
        \usepackage{fullpage}
    \else
        \documentclass[12pt,draftcls,onecolumn]{IEEEtran}
    \fi
\fi

\usepackage{calc,amsfonts,amssymb,amsmath,bm,url,color,graphicx,cite,epstopdf,nicefrac}
\usepackage{amsthm}
\usepackage{psfrag,subfigure,float}
\usepackage[algoruled,linesnumbered]{algorithm2e}
\usepackage{multirow}
\usepackage[normalem]{ulem}

\definecolor{orange}{RGB}{255,107,0}

\newtheorem{Lemma}{Lemma}

\newtheorem{Theorem}{Theorem}

\theoremstyle{definition}
\newtheorem{Def}{Definition}

\newcommand\bd{\ensuremath{{\rm bd}}}

\newcommand{\W}{\boldsymbol{W}}

\newcommand{\Q}{\boldsymbol{Q}}
\newcommand{\X}{\boldsymbol{X}}

\newcommand{\U}{\boldsymbol{U}}

\newcommand{\one}{\boldsymbol{1}}
\renewcommand{\H}{\boldsymbol{H}}
\newcommand{\M}{\boldsymbol{M}}
\newcommand{\A}{\boldsymbol{A}}

\newcommand{\x}{\boldsymbol{x}}

\renewcommand{\a}{\boldsymbol{a}}

\newcommand{\cone}[1]{\textup{cone}\{#1\}}
\newcommand{\T}{{\!\top\!}}
\DeclareMathOperator*{\argmin}{\arg\min}

\begin{document}

\newcommand{\papertitle}{
On Identifiability of Nonnegative Matrix Factorization
}

\newcommand{\paperabstract}{%
In this letter, we propose a new identification criterion that guarantees the recovery of the low-rank latent factors in the nonnegative matrix factorization (NMF) model, under mild conditions. Specifically, using the proposed criterion, it suffices to identify the latent factors if the rows of one factor are \emph{sufficiently scattered} over the nonnegative orthant, while no structural assumption is imposed on the other factor except being full-rank. This is by far the mildest condition under which the latent factors are provably identifiable from the NMF model.
}


\ifplainver

    \date{\today}

    \title{\papertitle}

    \author{
    Xiao Fu$^\ast$, Kejun Huang\footnote{The two authors contributed equally.} , and Nicholas D. Sidiropoulos
    \\ ~ \\
		Department of Electrical and Computer Engineering, University of Minnesota,\\
		Minneapolis, 55455, MN, United States\\
		Email: (xfu,huang663,nikos)@umn.edu
	  \\~	  
    }

    \maketitle

    \begin{abstract}
    \paperabstract
    \end{abstract}

\else
    \title{\papertitle}

    \ifconfver \else {\linespread{1.1} \rm \fi

\author{Xiao Fu$^\ast$, Kejun Huang$^\ast$, and Nicholas D. Sidiropoulos
	
\thanks{
This work is supported in part by National Science Foundation under Project NSF ECCS-1608961 and Project NSF IIS-1447788.
The authors were with the Department of ECE, University of Minnesota. X. Fu is now with the School of EECS, Oregon State University. N. D. Sidiropoulos is now with the Department of ECE, University of Virginia. emails: xiao.fu@oregonstate.edu, huang663@umn.edu, nikos@virginia.edu. 

$^\ast$ The two authors contributed equally.
}
}

    \maketitle

    \ifconfver \else
        \begin{center} \vspace*{-2\baselineskip}
        \end{center}
    \fi

    \begin{abstract}
    \paperabstract
    \end{abstract}

    \begin{IEEEkeywords}\vspace{-0.0cm}%
        Nonnegative matrix factorization, sufficiently scattered, convex analysis, identifiability
    \end{IEEEkeywords}

    \ifconfver \else \IEEEpeerreviewmaketitle} \fi

 \fi

\ifconfver \else
    \ifplainver \else
        \newpage
\fi \fi
\vspace{-.2cm}
\section{Introduction}
\vspace{-.1cm}
Nonnegative matrix factorization (NMF) \cite{lee1999learning,gillis2014and} aims to decompose a data matrix 
into low-rank latent factor matrices with nonnegativity constraints on (one or both of) the latent matrices. In other words, given a data matrix $\bm X\in\mathbb{R}^{M\times N}$ and a targeted rank $r$, NMF tries to find a factorization model ${\bm X}={\bm W}{\bm H}^\T$,
where ${\bm W}\in\mathbb{R}^{M\times r}$ and/or ${\bm H}\in\mathbb{R}^{N\times r}$ take only nonnegative values and $r\leq \min\{M,N\}$.

One notable trait of NMF is model identifiability -- the latent factors are uniquely identifiable under some conditions (up to some trivial ambiguities). Identifiability is critical in parameter estimation and model recovery.
In signal processing, many NMF-based approaches have therefore been proposed to handle problems such as blind source separation \cite{fu2015blind}, spectrum sensing \cite{fu2015power}, and hyperspectral unmixing \cite{Ma2013,fu2016robust}, where model identifiability plays an essential role.
In machine learning, identifiability of NMF is also considered essential for applications such as latent mixture model recovery \cite{anandkumar2012spectral}, topic mining \cite{arora2012practical}, and social network clustering \cite{mao2017mixed}, where model identifiability is entangled with interpretability of the results.

Despite the importance of identifiability in NMF, the analytical understanding of this aspect is still quite limited and 
many existing identifiability conditions for NMF are not satisfactory in some sense.
Donoho et al. \cite{donoho2003does}, Laurberg et al. \cite{laurberg2008theorems}, and Huang et al. \cite{huang2014non} have proven different sufficient conditions for identifiability of 
NMF, but these conditions all require that both of the generative factors $\W$ and $\H$ 
exhibit certain sparsity patterns or properties. The machine learning and remote sensing communities have proposed several factorization criteria and algorithms that have identifiability guarantees, but these methods heavily rely on the so-called \emph{separability condition} \cite{VMAX,Gillis2012,fu2015self,recht2012factoring,arora2012practical,elhamifar2013sparse,Esser2012}. The separability condition essentially assumes that there is a (scaled) permutation matrix in one of the two latent factors as a submatrix, which is clearly restrictive in practice. Recently, Fu et al. \cite{fu2015blind} and Lin et al. \cite{lin2014identifiability} proved that the so-called volume minimization (VolMin) criterion can identify $\W$ and $\H$ without any assumption on one factor (say, $\W$) except being full-rank when the other ($\H$) satisfies a condition which is much milder than separability. However, the caveat is that VolMin also requires that each row of the nonnegative factor sums up to one. This assumption implies loss of generality, and is not satisfied in many applications.

In this letter, we reveal a new identifiablity result for NMF, which is obtained from a delicate tweak of the VolMin identification criterion. Specifically, we `shift' the sum-to-one constraint on $\H$ from its rows to its columns. 
As a result, we show that this `constraint-altered VolMin criterion' identifies $\W$ and $\H$ with provable guarantees under conditions that are much more easily satisfied relative to VolMin.
This interesting tweak is seemingly slight yet the result is significant: putting sum-to-one constraints on the columns (instead of rows) of $\H$ is without loss of generality, since the bilinear model $\X=\W\H^\T$ can always be re-written as $\X=\W\bm D^{-1}(\H\bm D)^\T$, where $\bm D$ is a full-rank diagonal matrix satisfying $\bm D_{r,r}=1/\|\H_{:,r}\|_1$. Our new result is the only identifiability condition that 
does not assume {\em any other} structure beyond the target rank on $\W$ (e.g., zero pattern or nonnegativity) and has natural assumptions on $\H$ (relative to the restrictive row sum-to-one assumption as in VolMin). 

\vspace{-.3cm}
\section{Background}
\vspace{-.1cm}

To facilitate our discussion, let us formally define identifiability of constrained matrix factorization.

\vspace{-.15cm}
\begin{Def} (Identifiability)
	Consider a data matrix generated from the model $\X=\W_\natural\H_\natural^\T$, where $\W_\natural$ and $\H_\natural$ are the ground-truth factors. Let $(\W_\star,\H_\star)$ be an optimal solution from an identification criterion,
	\[  (\W_\star,\H_\star) = \argmin_{\X=\W\H^\T}~g(\W,\H). \]
	If $\W_\natural$ and/or $\H_\natural$ satisfy some condition such that for all $(\W_\star,\H_\star)$, we have that $\W_\star = \W_\natural{\bm \varPi}{\bm D}$ and $\H_\star = \H_\natural{\bm \varPi}{\bm D}^{-1}$, where ${\bm \varPi}$ is a permutation matrix and ${\bm D}$ is a full-rank diagonal matrix, then we say that the matrix factorization model is identifiable under that condition\footnote{Whereas identifiability is usually understood as a property of a given model that is independent of the identification criterion, NMF can be identifiable under a suitable identification criterion, but not under another, as we will soon see.}.
\end{Def}

\vspace{-.15cm}
For the `plain NMF' model~\cite{donoho2003does,lee1999learning,huang2014non,huang2014putting}, the identification criterion $g(\W,\H)$ is 1 (or, $\infty$) if ${\bm W}$ or ${\bm H}$ has a negative element, and 0 otherwise. 
Assuming that ${\X}$ can be perfectly factored under the postulated model, the above is equivalent to the popular least-squares NMF formulation:
\begin{align}\label{eq:nmf_fit}
	\min_{\W\geq{\bm 0},~{\bm H}\geq {\bm 0},}&~\left\| \X - \W\H^\T\right\|_F^2.
\end{align}
Several sufficient conditions for identifiability of \eqref{eq:nmf_fit} have been proposed. Early results in \cite{donoho2003does,laurberg2008theorems} require that one factor (say, ${\H}$) satisfies the so-called the separability condition:
\vspace{-.15cm}
\begin{Def}[Separability]
A matrix $\H\in\mathbb{R}_+^{N\times r}$ is separable if for every $k=1,...,r$, there exists a row index $n_k$ such that 
$\H_{n_k,:} = \alpha_k\bm{e}_k^\T$, where $\alpha_k>0$ is a scalar and $\bm{e}_k$ is the $k$th coordinate vector in $\mathbb{R}^r$.
\end{Def}
\vspace{-.15cm}
With the separability assumption, the works in \cite{donoho2003does,laurberg2008theorems} first revealed the reason behind the success of NMF in many applications -- NMF is unique under some conditions.
The downside is that separability is easily violated in practice -- see discussions in \cite{Ma2013}.
In addition, the conditions in \cite{donoho2003does,laurberg2008theorems} also need that $\W$ to exhibit a certain zero pattern on top of $\H$ satisfying separability. This is also considered restrictive in practice -- e.g., in hyperspectral unmixing, ${\W}_{:,r}$'s are spectral signatures, which are always dense. 
The remote sensing and machine learning communities have come up with many different separability-based identification methods without assuming zero patterns on $\W$, e.g., the volume maximization (VolMax) criterion \cite{VMAX,arora2012practical} and self-dictionary sparse regression \cite{fu2015self,fu2015robust,gillis2013robustness,recht2012factoring,arora2012practical}, respectively. 
However, the separability condition was not relaxed in those works.

\begin{figure}[t]
	\centering
	\includegraphics[width=0.75\linewidth]{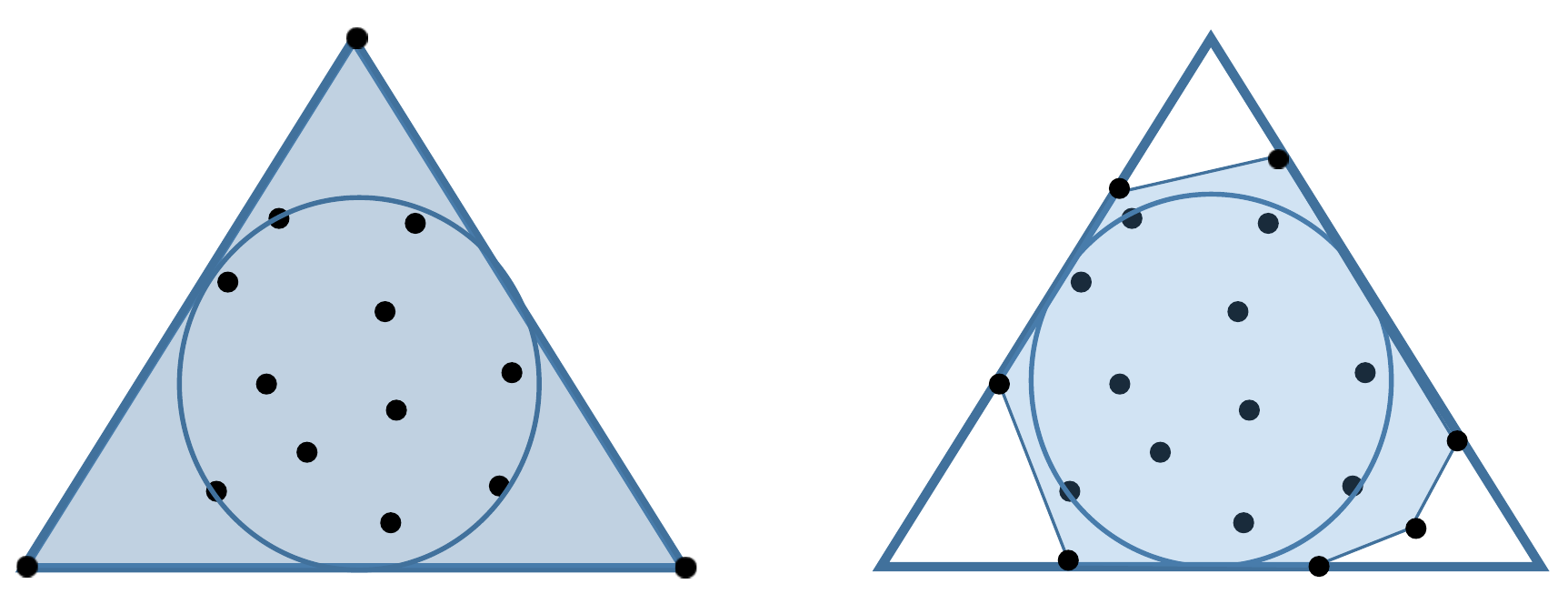}
	\vspace{-8pt}
	\caption{Illustration of the separability (left) and sufficiently scattered (right) conditions by assuming that the viewer stands in the nonnegative orthant and faces the origin. The dots are rows of $\H$; the triangle is the nonnegative orthant; the circle is ${\cal C}$; the shaded region is $\cone{\H^\T}$. Clearly, separability is special case of the sufficiently scattered condition.}\label{fig:cond}
	\vspace{-14pt}
\end{figure}

The stringent separability condition was considerably relaxed by Huang et al. \cite{huang2014non} based on a so-called \emph{sufficiently scattered} condition from a geometric interpretation of NMF. 
\vspace{-.15cm}
\begin{Def}[Sufficiently Scattered]\label{def:suf}
	A matrix $\H\in\mathbb{R}_+^{N\times r}$ is sufficiently scattered if 1) $\cone{\H^\T} \supseteq {\cal C}$,
		2) $\cone{\H^\T}^* \cap \bd{\cal C}^* = \{\lambda\bm e_k~|~\lambda\geq 0,k=1,...,r \}$,
	where
	$\mathcal{C} = \{\x | \x^\T\mathbf{1} \geq \sqrt{r-1}\|\x\|_2\}$,
	$\mathcal{C}^* = \{\x | \x^\T\mathbf{1} \geq\|\x\|_2\}$, $\cone{\H^\T}=\{{\bm x}|{\bm x}=\H^\T\bm \theta,~\forall \bm \theta\geq{\bm 0}\}$ and $\cone{\H^\T}^\ast = \{{\bm y}|\x^\T \bm y \geq{\bm 0}, ~\forall \x\in\cone{\H^\T}\}$ are the conic hull of $\H^\T$ and its dual cone, respectively,
	and ${\rm bd}$ denotes the boundary of a set.
\end{Def}
\vspace{-.15cm}
The main result in \cite{huang2014non} is that if both $\W$ and $\H$ satisfy the sufficiently scattered condition, then the criterion in \eqref{eq:nmf_fit} has identifiability. This is a notable result since it was the first provable result in which separability was relaxed for \emph{both} $\W$ and $\H$. 
The sufficiently scattered condition essentially means that $\cone{\H^\T}$ contains ${\cal C}$
as its subset, which is much more relaxed than separability that needs $\cone{\H^\T}$ to contain the entire nonnegative orthant; see Fig.~\ref{fig:cond}.

\begin{figure}[t]
	\centering
	\includegraphics[width=0.55\linewidth]{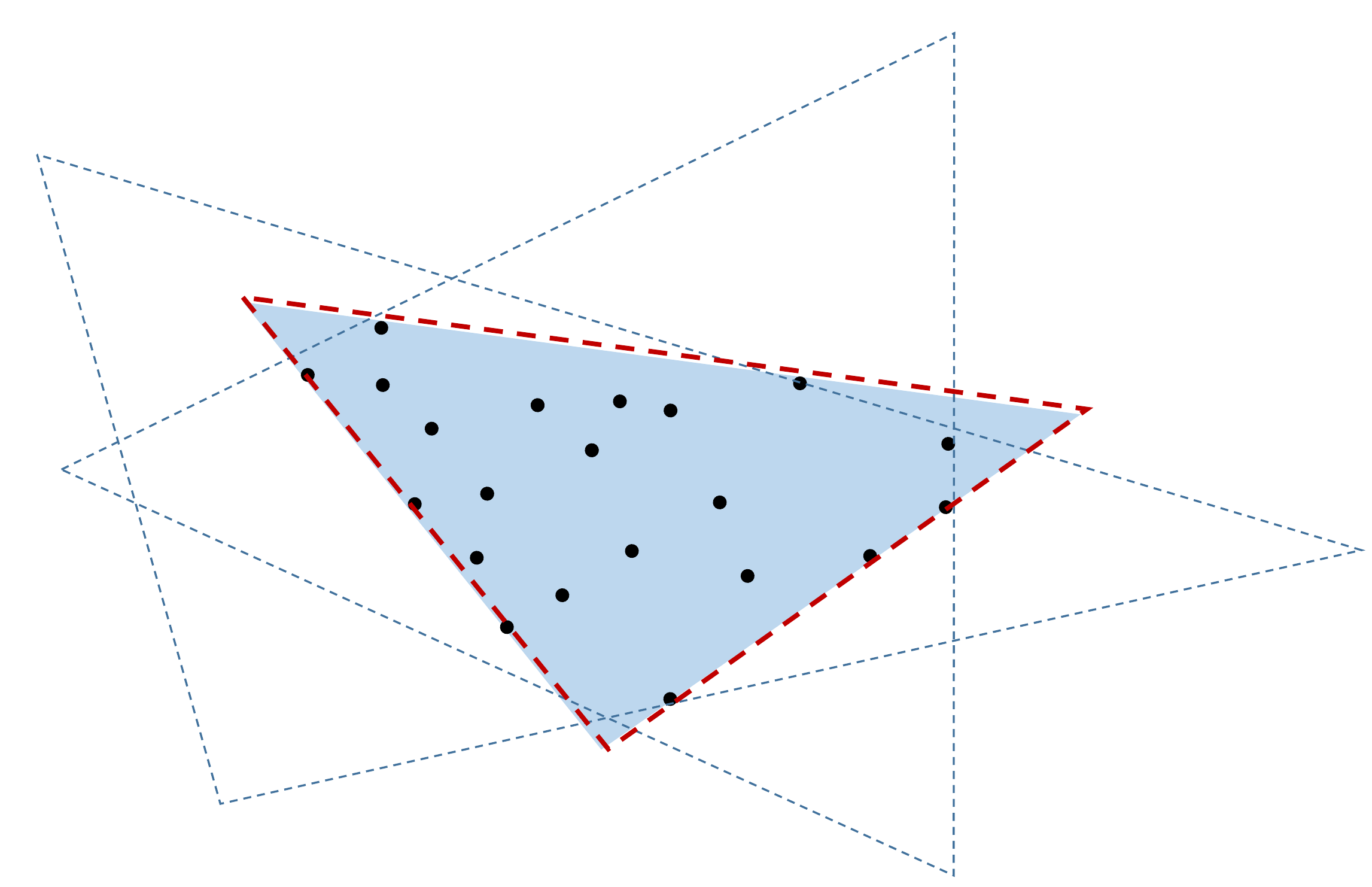}
		\vspace{-8pt}
	\caption{The intuition of VolMin. The shaded region is ${\sf conv}\{\W_{:,1},\ldots,\W_{:,r}\}$; the dots are $\X_{:,n}$'s; the dash lines are enclosing convex hulls; the bold dashed lines comprise the minimum-volume data-enclosing convex hull.}\label{fig:volmin}
	\vspace{-.65cm}
\end{figure}

On the other hand, the zero-pattern assumption on $\W$ and $\H$ are still needed in \cite{huang2014non}. 
Another line of work removed the zero pattern assumption from one factor (say, $\W$) by using a different identification criterion \cite{fu2015blind,lin2014identifiability}:
\begin{subequations}\label{eq:volmin}
	\begin{align}
	\min_{\W\in\mathbb{R}^{M\times r},\H\in\mathbb{R}^{N\times r}}&~\det\left({\bm W}^\T{\bm W}\right)\\
	{\rm s.t.~~~~~~~~}&~\X = \W\H^\T,\\
	&~{\bm H}{\bm 1}={\bm 1},~{\bm H}\geq {\bm 0}, \label{eq:Hvolmin}
	\end{align}
\end{subequations}
where $\bm 1$ is an all-one vector with proper length.
Criterion~\eqref{eq:volmin} aims at finding the minimum-volume (measured by determinant) data-enclosing convex hull (or simplex). The main result in \cite{fu2015blind} is that if the ground-truth $\H\in\{{\bm Y}\in\mathbb{R}^{N\times r}|{\bm Y}{\bm 1}={\bm 1},{\bm Y}\geq{\bm 0}\}$ and $\bm H$ is sufficiently scattered, then, the volume minimization (VolMin) criterion identifies the ground-truth $\W$ and ${\bm H}$. 
This very intuitive result is illustrated in Fig.~\ref{fig:volmin}: if $\H$ is sufficiently scattered in the nonnegative orthant, $\X_{:,n}$'s are sufficiently spread in the \emph{convex hull}
\footnote{The convex hull of $\W$ is defined as $\textup{conv}\{\W_{:,1},\ldots,\W_{:,r}\}=\{{\bm x}|\bm x={\bm W}{\bm \theta},\forall \bm \theta\geq {\bm 0},{\bm 1}^\T{\bm \theta}=1\}$.} spanned by the columns of $\W$. Then, finding the minimum-volume data-enclosing convex hull recovers the ground-truth $\W$.
This result resolves the long-standing \emph{Craig's conjecture} in remote sensing~\cite{craig1994minimum} proposed in the 1990s.

The VolMin identifiability condition is intriguing since it completely sets $\W$ free -- there is no assumption on the ground-truth $\W$ except for being full-column rank, and it has a very mild assumption on $\H$. There is a caveat, however: The VolMin criterion needs an extra condition on the ground-truth $\H$, namely $\H{\bm 1}={\bm 1}$, so that the columns of $\X$ all live in the convex hull (not \emph{conic hull} as in the general NMF case) spanned by the columns of $\W$ -- otherwise, the geometric intuition of VolMin in Fig.~\ref{fig:volmin} does not make sense. 
Many NMF problem instances stemming from applications do not naturally satisfy this assumption. The common trick is to normalize the columns of $\X$ using their $\ell_1$-norms~\cite{Gillis2012} so that an equivalent model with this sum-to-one assumption holding is enforced -- but normalization only works when the ground-truth $\W$ is also nonnegative.
This raises a natural question: can we essentially keep the advantages of VolMin identifiability (namely, no structural assumption on $\W$ (other than low-rank) and no separability requirement on $\H$) without assuming sum-to-one on the rows of the ground-truth $\H$? 

\vspace{-.15cm}
\section{Main Result}
\vspace{-.15cm}
Our main result in this letter fixes the issues with the VolMin identifiability. Specifically, we show that, with a careful and delicate tweak to the VolMin criterion, one can identify the model $\X=\W{\H}^\T$ without assuming the sum-to-one condition on the rows of $\H$:

\vspace{-.25cm}
\begin{Theorem}
Assume that $\X=\W_\natural\H_\natural^\T$ where $\W_\natural\in\mathbb{R}^{M\times r}$ and $\H_\natural\in\mathbb{R}^{N\times r}$ and that ${\rm rank}(\X)={\rm rank}(\W_\natural)=r$. Also, assume that $\H_\natural$ is sufficiently scattered. Let $(\W_\star,\H_\star)$ be the optimal solution of the following identification criterion:
\begin{subequations}\label{eq:new}
	\begin{align}
	\min_{\W\in\mathbb{R}^{M\times r},\H\in\mathbb{R}^{N\times r}}&~\det\left({\bm W}^\T{\bm W}\right)\\
	{\rm s.t.~~~~~~~~}&~\X = \W\H^\T,\\
	&~{\bm H}^\T{\bm 1}={\bm 1},~{\bm H}\geq {\bm 0}. \label{eq:Hnew}
	\end{align}
\end{subequations}
Then, $\W_\star = \W_\natural{\bm \Pi}{\bm D}$ and $\H_\star = \H_\natural{\bm \Pi}{\bm D}^{-1}$ must hold, where ${\bm \Pi}$ and ${\bm D}$ denotes a permutation matrix and a full-rank diagonal matrix, respectively,	
\end{Theorem}
\vspace{-.2cm}

At first glance, the identification criterion in \eqref{eq:new} looks similar to VolMin in \eqref{eq:volmin}. The difference lies between \eqref{eq:Hvolmin} and \eqref{eq:Hnew}. In \eqref{eq:Hnew}, we `shift' the sum-to-one condition to the \emph{columns} of $\H$, rather than enforcing it on the rows of $\H$. This simple modification makes a big difference in terms of generality: Enforcing columns of $\H$ to be sum-to-one entails no loss in generality, since in bilinear factorization models like $\X=\W\H^\T$ there is always an intrinsic scaling ambiguity of the columns. In other words, one can always assume the columns of $\H$ are scaled by a diagonal matrix and then counter scale the corresponding columns of $\W$, which will not affect the factorization model; i.e., $\X=(\W\bm D^{-1})(\H\bm D)^\T$ still holds. Therefore, there is no need for data normalization to enforce this constraint, as opposed to the  VolMin case. In fact, the identifiability of \eqref{eq:new} holds for $\H^\T{\bm 1}=\rho{\bm 1}$ for any $\rho>0$ -- we use $\rho=1$ only for notational simplicity.

We should mention that avoiding normalization is a significant advantage in practice even when $\W\geq{\bm 0}$ holds, especially when there is noise -- since normalization may amplify noise. It was also reported in the literature that normalization degrades performance of text mining significantly since it usually worsens the conditioning of the data matrix \cite{kumar2012fast}. In addition, as mentioned, in applications where $\W$ naturally contains negative elements (e.g., channel identification in MIMO communications), even normalization cannot enforce the VolMin model.

It is worth noting that the criterion in Theorem 1 has by far the most relaxed identifiability conditions for nonnegative matrix factorization. A detailed comparison of different NMF conditions are listed in Table~\ref{tab:compare}, where one can see that Criterion \eqref{eq:new} works under the mildest conditions on both $\H$ and $\W$. Specifically, compared to plain NMF, the new criterion does not assume any structure on $\W$; compared to VolMin, it does not need the sum-to-one assumption on the rows of $\H$ or nonnegativity of $\W$;
it also does not need separability, which is inherited from the advantage of VolMin.

%

In the next section, we will show the proof of Theorem~1.
We should remark that the although it seems that shifting the sum-to-one constraint to the columns of $\H$ is a `small' modification to VolMin, the result in Theorem 1 was not obvious at all before we proved it: by this modification, the clear geometric intuition of VolMin no longer holds -- the objective in \eqref{eq:new} no longer corresponds to the volume of a data-enclosing convex hull and has no geometric interpretation any more.
Indeed, our proof for the new criterion is purely algebraic rather than geometric.

\begin{table}[t]
	\begin{center}
		\centering

	\caption{Different assumptions on $\W$ and $\H$ for identifiability of NMF.}
	\vspace{-7pt}
		\resizebox{\linewidth}{!}{
    \begin{tabular}{|c|c|c|c|c|c|}
    	\hline
    	& plain \cite{huang2014non} & Self-dict \cite{fu2015self,gillis2013robustness,recht2012factoring} & VolMax \cite{VMAX,arora2012practical} & VolMin \cite{fu2015blind,lin2014identifiability} & Proposed \\
    	\hline
    	\multirow{2}[2]{*}{$\W$} & \multirow{2}[2]{*}{NN, Suff.} & NN, Full-rank & NN, Full-rank & NN, Full-rank & \multirow{2}[2]{*}{Full-rank} \\
    	&       & (Full-rank) & ( Full-rank) & (Full-rank) & \\
    	\hline
    	\multirow{2}[2]{*}{$\H$} & \multirow{2}[2]{*}{NN. Suff. } & NN. Sep. & NN. Sep & NN., Suff. & \multirow{2}[2]{*}{NN.  Suff.} \\
    	&       & (NN., Sep., row sto) & (NN., Sep., row sto) & (NN., Suff., row sto) & \\
    	\hline
    \end{tabular}

}%
	\label{tab:compare}%
	\smallskip
	\end{center}	
	\vspace{-.15cm}
	Note: `NN' means nonnegativity, `Sep.' means separability, `Suff.' denotes the sufficiently scattered condition, and `sto' denotes sum-to-one. The conditions in `$(\cdot)$' give an alternative set of conditions for the corresponding approach.
	\vspace{-.55cm}
\end{table}%

\vspace{-.25cm}
\section{Proof of Theorem 1}
\vspace{-.15cm}
The major insights of the proof are evolved from the VolMin work of the authors and variants \cite{fu2015blind,huangprincipled,huafusid2016nips}, with proper modifications to show Theorem 1.
To proceed, let us first introduce the following classic lemma in convex analysis:
\vspace{-.3cm}
\begin{Lemma} \cite{rockafellar1997convex} \label{lem:dualcone}
	If ${\cal K}_1$ and ${\cal K}_2$ are convex cones and ${\cal K}_1\subseteq {\cal K}_2$, then,
${\cal K}_2^\ast \subseteq {\cal K}_1^\ast,$
	where ${\cal K}_1^\ast$ and ${\cal K}_2^\ast$ denote the dual cones of ${\cal K}_1$ and ${\cal K}_2$, respectively.
\end{Lemma}
\vspace{-.3cm}


Our purpose is to show that the optimization criterion in \eqref{eq:new} outputs $\W_\star$ and $\H_\star$ that are the column-scaled and permuted versions of the ground-truth $\W_\natural$ and $\H_\natural$. To this end, let us denote
$(\widehat{\W}\in\mathbb{R}^{M\times r},\widehat{\H}\in\mathbb{R}^{N\times r})$ as a feasible solution of Problem~\eqref{eq:new} that satisfies the constraints in \eqref{eq:new}, i.e.,
\begin{align}
	&\X = \widehat{\W}\widehat{\H}^{\,\T}, \quad\widehat{\H}^{\,\T}{\bm 1}={\bm 1},~\widehat{\H}\geq{\bm 0}. \label{eq:const2}
\end{align}
Note that $\X=\W_\natural\H_\natural^\T$ and that $\W_\natural$ has full-column rank.
In addition, since $\H_\natural$ is sufficiently scattered, ${\rm rank}(\H_\natural)=r$ also holds \cite[Lemma1]{huafusid2016nips}.
Consequently, there exists an invertible $\A\in\mathbb{R}^{r\times r}$ such that 
\begin{equation}\label{eq:rotation}
\widehat{\bm H}=\H_\natural\A, \widehat{\W}={\bm W}_\natural{\bm A}^{-\T}. 
\end{equation}
This is because $\widehat{\W}$ and $\widehat{\H}$ has to have full column-rank and thus $\H_\natural$ and $\widehat{\H}$ span the same subspace. Otherwise, ${\rm rank}({\X})=r$ cannot hold.
Since \eqref{eq:const2} holds, one can see that
\begin{equation}
  \widehat{\H}^{\,\T}{\bm 1} =  \A^\T\H_\natural^\T\one=\A^\T\one=\one.
\end{equation}
By \eqref{eq:const2}, we also have
${\H}_\natural\A \geq {\bm 0}.$
By the definition of a dual cone, $\H_\natural\A\geq 0$ means that $\a_i\in\cone{\H^\T_\natural}^*$, where $\a_i$ is the $i$-column of $\A$, for all $i=1,...,r$. Because $\H_\natural$ is sufficiently scattered, we have that 
${\cal C}\subseteq \cone{\H_\natural}$
which, together with Lemma~1, leads to
$\cone{\H^\T_\natural}\subseteq\mathcal{C}^*$. This further implies that $\a_i \in \mathcal{C}^*$, which means
$\|\a_i\|_2 \leq \one^\T\a_i,$
by the definition of ${\cal C}^\ast$.
Then we have the following chain
\begin{subequations}\label{eq:proofvolcol}
	\begin{align}
	|\det(\A)| & \leq \prod_{i=1}^{k}\|\a_i\|_2 \label{eq:proofvolcol1}\\
	& \leq \prod_{i=1}^{k}\one^\T\a_i \label{eq:proofvolcol2} \\
	& = 1,
	\end{align}
\end{subequations}
where (\ref{eq:proofvolcol1}) is Hadamard's inequality, and (\ref{eq:proofvolcol2}) is due to $\a_i\!\in\!\mathcal{C}^*$.

Now, suppose the equality is attained, i.e., $|\det(\A)|=1$, then all the inequalities in (\ref{eq:proofvolcol}) hold as equality, and specifically (\ref{eq:proofvolcol2}) means that the columns of $\A$ lie on the boundary of $\mathcal{C}^*$. Recall that $\a_i\in\cone{\H_\natural^\T}^*$, and $\H_\natural$ being sufficiently scattered, according to the second requirement in Definition~\ref{def:suf}, shows that
$\cone{\H^\T_\natural}^* \cap \bd{\mathcal{C}^*} = \{\lambda\bm e_k~|~\lambda\geq 0,k=1,...,r \},$
therefore $\a_i$'s can only be the $\bm e_k$'s. In other words, $\A$ can only be a permutation matrix.

Suppose that an optimal solution ${\H}_\star$ of (\ref{eq:new}) is not a column permutation of $\H_\natural$. Since $\W_\natural$ and $\H_\natural$ are clearly feasible for (\ref{eq:new}), this means that $\det(\W_\star^\T\W_\star) \leq \det(\W_\natural^\T\W_\natural)$.
We also know that for every feasible solution, including $\W_\star$ and $\H_\star$, Eq.~\eqref{eq:rotation} holds,
which means we have
$\H_\star=\H_\natural\A$ and $\W_\star={\bm W}_\natural{\bm A}^{-\T}$ hold for a certain invertible $\A\in\mathbb{R}^{r\times r}$.
Since $\H_\natural$ is sufficiently scattered, according to \eqref{eq:proofvolcol2}, and our assumption that $\A$ is not a permutation matrix, we have
$	|\det(\A)|<1.$
However, the optimal objective of \eqref{eq:new} is
\begin{align*}
	\det(\W_\star^\T\W_\star) &= \det(\A^{-1}\W_\natural^\T\W_\natural\A^{-\T}) \\
	&= \det(\A^{-1}) \det(\W_\natural^\T\W_\natural) \det(\A^{-\T}) \\
	&= |\det(\A)|^{-2} \det(\W_\natural^\T\W_\natural) \\
	&> \det(\W_\natural^\T\W_\natural),
\end{align*}
which contradicts our first assumption that $(\W_\star, \H_\star)$ is an optimal solution for \eqref{eq:new}. Therefore, $\H_\star$ must be a column permutation of $\H_\natural$. \hfill {\bf Q.E.D.}

As a remark, the proof of Theorem~1 follows the same rationale of that of the VolMin identifiability as in \cite{fu2015blind}.
The critical change is that we have made use of the relationship between sufficiently scattered $\H$ and the inequality in \eqref{eq:proofvolcol} here. This inequality appeared in \cite{huangprincipled,huafusid2016nips} but was not related to the bilinear matrix factorization criterion in \eqref{eq:new} -- which might be by far the most important application of this inequality. The interesting and surprising point is that, by this simple yet delicate tweak , the identifiability criterion can cover a substantially wider range of applications which naturally involve 
$\W$'s that are not nonnegative.

\vspace{-.1cm}
\section{Validation and Discussion}
\vspace{-.1cm}
The identification criterion in \eqref{eq:new} is a nonconvex optimization problem.
In particular, the bilinear constraint $\X=\W\H^\T$ is not easy to handle. 
However, the existing work-arounds for handling VolMin can all be employed to deal with Problem~\eqref{eq:new}.
One popular method for VolMin is to first take the singular value decomposition (SVD) of the data ${\X}=\bm U\bm \Sigma \bm V^\T$,
where $\U\in\mathbb{R}^{M\times r}$, $\bm \Sigma \in\mathbb{R}^{r\times r}$ and $\bm V\in\mathbb{R}^{N\times r}$. Then, $\bm V^\T = \widetilde{\bm W}\H^\T$ holds where $\widetilde{\W}\in\mathbb{R}^{r\times r}$ is invertible, because $\bm V$ and $\bm H$ span the same range space.
One can use \eqref{eq:new} to identify $\H$ from the data model $\widetilde{\X}=\bm V^\T=\widetilde{\W}\H^\T$.
Since $\widetilde{\W}$ is square and nonsingular, it has an inverse $\Q=\widetilde{\W}^{-1}$. The identification criterion in \eqref{eq:new} can be recast as
$	\max_{\Q\in\mathbb{R}^{r\times r}}~|\det\left({\bm Q}\right)|,~
	{\rm s.t.}
	~\Q\widetilde{\X} {\bm 1}={\bm 1},~\Q\widetilde{\X} \geq {\bm 0}. $
This reformulated problem is much more handy from an optimization point of view. 
To be specific, one can fix all the columns in $\bm Q$ except one, e.g., $\bm q_i$.
Then the optimization w.r.t. $\bm q_i$ is a linear function, i.e., $\det(\bm Q)=\sum_{i=1}^{r}(-1)^{i+k}\bm Q_{k,i}\det(\overline{\Q}_{k,i})={\bm p}^\T{\bm q}_i$, where ${\bm p}=[p_1,\ldots,p_r]^\T$, $p_k=(-1)^{i+k}\det(\overline{\Q}_{k,i}),~\forall~k=1,...,r$, and $\overline{\Q}_{k,i}$ is a submatrix of $\bm Q$ without the $k$th row and $i$th column of $\bm Q$. Maximizing $|\bm p^\T\bm q_i|$ subject to linear constraints can be solved via maximizing both $\bm p^\T\bm q_i$ and $-\bm p^\T\bm q_i$, followed by picking the solution that gives larger absolute objective. Then, cyclically updating the columns of $\M$ results in an alternating optimization (AO) algorithm.
Similar SVD and AO based solvers were proposed to handle VolMin and its variants in \cite{MVES,huangprincipled,huafusid2016nips}, and empirically good results have been observed.
Note that the AO procedure is not the only possible solver here. 
When the data is very noisy, one can reformulate the problem in \eqref{eq:new} as
\vspace{-.1cm}
$\min_{{\bm W},{\bm H}^\T{\bm 1}={\bm 1},{\bm H}\geq{\bm 0}}~\left\|\X-\W\H^\T\right\|_F^2+\lambda\det(\W^\T\W),$
where $\lambda>0$ balances the determinant term and the data fidelity. 
Many algorithms for regularized NMF can be employed and modified to handle the above.

An illustrative simulation is shown in Table~\ref{tab:result} to showcase the soundness of the theorem.
In this simulation, we generate $\X=\W_\natural\H_\natural^\T$ with $r=5,10$ and $M=N=200$. We tested several cases.
1) $\W_\natural\geq{\bm 0}$, $\H_\natural\geq{\bm 0}$, and both $\W_\natural$ and $\H_\natural$ are sufficiently scattered; 2) $\W_\natural\geq{\bm 0}$, $\H_\natural\geq{\bm 0}$, and $\H_\natural$ is sufficiently scattered but $\W_\natural$ is completely dense; 3) $\W_\natural$ follows the i.i.d. normal distribution, and $\H_\natural\geq {\bm 0}$ is sufficiently scattered.
We generate sufficiently scattered factors following \cite{kim2008nonnegative} -- i.e., we generate the elements of a factor following the unifom distribution between zero and one and zero out $35\%$ of its elements, randomly. This way, the obtained factor is empirically sufficiently scattered with an overwhelming probability. We employ the algorithm for fitting-based NMF in \cite{huang2014putting}, the VolMin algorithm in \cite{bioucas2009variable}, and the described algorithm to handle the new criterion, respectively. We measure the performance of different approaches by measuring the mean-squared-error (MSE) of the estimated $\widehat{\H}$, which is defined as
${\rm MSE}=\min_{\bm{\pi} \in \Pi}\frac{1}{r} \sum_{k=1}^r \left\| \nicefrac{ {\H_\natural}_{:,k }}{ \| {\bm H_\natural}_{:,k} \|_2 } - \nicefrac{ \widehat{\H}_{:,{\pi_k}} }{ \| \widehat{\H}_{:,{\pi_k}} \|_2 }    \right\|_2^2,$ 
where $\Pi$ is the set of all permutations of $\{ 1,2,\ldots,r \}$.
The results are obtained by averaging 50 random trials.

Table~\ref{tab:result} matches our theoretical analysis.
All the algorithms work very well on case 1, where both $\W_\natural$ and $\H_\natural$ are sparse (sp.) and sufficiently scattered.
In case 2, since $\W$ is nonegative yet dense (den.), plain NMF fails as expected, but VolMin still works, since normalization can help enforce its model when $\W\geq{\bm 0}$.
In case 3, when $\W$ follows the i.i.d. normal distribution, VolMin fails since normalization does not help -- while the proposed method still works perfectly.

\begin{table}[t]
	\centering
	\caption{MSEs of the estimated $\widehat{\H}$.}\vspace{-7pt}
	\resizebox{\linewidth}{!}{
    \begin{tabular}{|c|c|c|c|}
    	\hline
    	\multirow{2}[4]{*}{Method} & \multicolumn{3}{c|}{MSE of $\H$} \\
    	\cline{2-4}          & case 1 (sp. $\W$)  & case 2 (den. $\W$) & case 3 (Gauss. $\W$) \\
    	\hline
    	{Plain} ($r=5$) & 5.49E-05 & 0.0147 & 0.7468 \\
    	VolMin ($r=5$) & 1.36E-08 & 7.31E-10 & 1.0406 \\
    	Proposed ($r=5$) & 7.32E-18 & 7.78E-18 & 8.44E-18 \\
    	\hline
    	{Plain} ($r=10$) & 4.82E-04 & 0.0403 & 0.8003 \\
    	VolMin ($r=10$) & 8.64E-09 & 8.66E-09 & 1.2017 \\
    	Proposed ($r=10$) & 6.54E-18 & 5.02E-18 & 6.38E-18 \\
    	\hline
    \end{tabular}
}
	\label{tab:result}%
	\vspace{-16pt}
\end{table}%

To conclude, in this letter we discussed the identifiability issues with the current NMF approaches. We proposed a new NMF identification criterion that is a simple yet careful tweak of the existing volume minimization criterion. We show that, by slightly modifying the constraints of VolMin, the identifiability of the proposed criterion holds under the same sufficiently scattered condition in VolMin, but the modified criterion covers a much wider range of applications including the cases where one factor is not nonnegative. This new criterion offers identifiability to the largest variety of cases amongst the known results.


\clearpage
\bibliographystyle{IEEEtran}
\bibliography{refs_aug2015,refs_topic_model}
\end{document}